\documentclass[preprint,12pt]{elsarticle}

\usepackage{amssymb}

\usepackage[utf8]{inputenc} 
\usepackage{booktabs}       
\usepackage{amsfonts}       
\usepackage{nicefrac}       
\usepackage{microtype}      
\usepackage{amsmath}
\usepackage{amsthm}
\usepackage{amssymb}
\usepackage{algorithm}
\usepackage{algorithmic}
\usepackage{subfigure}
\usepackage{array}
\usepackage[switch]{lineno}
\usepackage{multirow}
\usepackage{color, xcolor}
\usepackage{url}
\usepackage{ulem}

\newtheorem{remark}{Remark}

\journal{Neural Networks}

\begin{document}

\begin{frontmatter}



\author[1,2]{Xunyu Zhu}
\ead{zhuxunyu@iie.ac.cn}

\author[1,2]{Jian Li\corref{cor1}}
\ead{lijian9026@iie.ac.cn}

\author[3]{Yong Liu}
\ead{liuyonggsai@ruc.edu.cn}

\author[1,2]{Weiping Wang}
\ead{wangweiping@iie.ac.cn}

\cortext[cor1]{Corresponding author}

\affiliation[1]{Institute of Information Engineering, Chinese Academy of Sciences.}

\affiliation[2]{School of Cyber Security, University of Chinese Academy of Sciences.}

\affiliation[3]{Gaoling School of Artificial Intelligence, Renmin University of China.}

\title{Improving Differentiable Architecture Search via Self-Distillation}

\begin{abstract}

 Differentiable Architecture Search (DARTS) is a simple yet efficient Neural Architecture Search (NAS) method. During the search stage, DARTS trains a supernet by jointly optimizing architecture parameters and network parameters. During the evaluation stage, DARTS discretizes the supernet to derive the optimal architecture based on architecture parameters. However, recent research has shown that during the training process, the supernet tends to converge towards sharp minima rather than flat minima. This is evidenced by the higher sharpness of the loss landscape of the supernet, which ultimately leads to a performance gap between the supernet and the optimal architecture. In this paper, we propose Self-Distillation Differentiable Neural Architecture Search (SD-DARTS) to alleviate the discretization gap. We utilize self-distillation to distill knowledge from previous steps of the supernet to guide its training in the current step, effectively reducing the sharpness of the supernet's loss and bridging the performance gap between the supernet and the optimal architecture. Furthermore, we introduce the concept of voting teachers, where multiple previous supernets are selected as teachers, and their output probabilities are aggregated through voting to obtain the final teacher prediction. Experimental results on real datasets demonstrate the advantages of our novel self-distillation-based NAS method compared to state-of-the-art alternatives. 
\end{abstract}



\begin{keyword}
neural architecture search, neural networks, flatness, knowledge distillation, sharpness-aware minimization 
\end{keyword}

\end{frontmatter}


\section{Introduction}


Deep learning has gained significant popularity across various domains due to its ability to automate feature learning and achieve impressive performance. It has been successfully applied in areas such as image classification \cite{deng2009imagenet}, object tracking \cite{yilmaz2006object}, image processing \cite{van1996new}, and text detection \cite{li2000automatic}. To enhance the efficiency of deep learning, several outstanding architectures have been developed by researchers and professors, including VGG \cite{simonyan2014very}, ResNet \cite{he2016deep}, Transformer \cite{vaswani2017attention}, Bert \cite{devlin2018bert}, and GPT \cite{brown2020language}. These architectures have significantly contributed to improving the performance of deep learning models.

However, designing high-quality architectures manually can be a laborious and time-consuming process. The cost of manual architecture design is often high, requiring extensive expertise and trial-and-error experimentation. As a result, \textit{Neural Architecture Search} (NAS) has gained popularity in academia and industry as it enables machines to automatically discover architectures with good performance. Initially, NAS approaches such as Reinforcement Learning \cite{zoph2016neural} and Evolutionary Algorithms \cite{real2019regularized, ostad2021subsurface} are introduced, which involve sampling and evaluating individual architectures by training them from scratch. However, these methods face challenges in dealing with the exponentially increasing search space, resulting in extensive computational resources being required to find the optimal architecture. NASNet \cite{zoph2018learning, ostad2017artificial} introduces a cell-based approach, where individual cells are searched and then stacked together to form the final network. Another approach, ENAS \cite{pham2018efficient}, introduces the concept of a supernet, where a single network is trained with shared weights that serve as a proxy for evaluating the performance of individual subnetworks. These methods significantly reduce the computational cost associated with NAS. However, NAS is still considered a discrete optimization problem. In response, gradient-based methods such as SPOS \cite{DBLP:conf/eccv/GuoZMHLWS20} and NAO \cite{luo2018neural} are proposed. These methods relaxes the NAS problem as a continuous optimization problem and utilized gradient-based optimizers to explore the continuous search space. Among these methods, DARTS stands out as one of the most successful gradient-based approaches, offering high search efficiency.

\textit{Differentiable Architecture Search} (DARTS) revolutionizes NAS by formulating it as a continuous optimization problem and leveraging gradient-based optimization techniques. DARTS achieves low search cost, requiring only 0.3 GPU days to discover high-quality architectures. To further improve efficiency, PC-DARTS \cite{xu2019pc} introduces partial channel connections and edge normalization to reduce memory consumption and computational overhead. SGAS \cite{li2020sgas} introduces sequential greedy architecture search to address the degenerate search-evaluation correlation problem. During the evaluation stage of DARTS, the supernet is discretized to determine the optimal architecture. However, despite consistently reducing the validation error of the supernet and the optimal architecture, a performance gap remains during evaluation. Several papers \cite{zela2019understanding, chen2020stabilizing} have discussed this issue and attributed the performance gap to the geometry of the supernet's loss landscape. SDARTS \cite{chen2020stabilizing} and R-DARTS \cite{zela2019understanding} identifiy the sharpness of the validation loss landscape in DARTS as the cause of the performance gap. SDARTS introduces the generation of weight perturbations to improve the supernet's training process, but this approach incured a significant computational cost. R-DARTS uses the dominant eigenvalue of the Hessian norm as an indicator, employing early stopping when the indicator reaches a threshold. However, R-DARTS prevents DARTS from exploring architectures with potentially better performance. Alleviating the discretization gap by flattening the supernet's loss landscape remains an open question that warrants further exploration. 

\begin{figure*}[]
	\vskip 0.2in
	\begin{center}
		\centerline{\includegraphics[width=\linewidth]{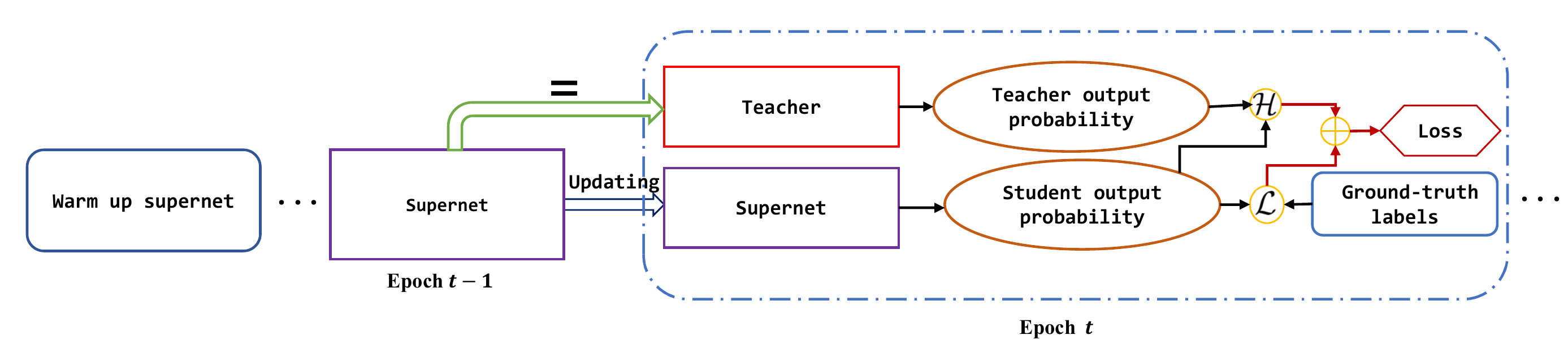}}
		\caption{The training stage of SD-DARTS involves utilizing the supernet from epoch $t-1$ as the teacher for epoch $t$. The correlation between the teacher's output probability and the supernet's output probability is calculated, serving as a regularization term to guide the supernet's training.}
		\label{sd-darts}
	\end{center}
	\vskip -0.2in
\end{figure*}

Our paper introduces a novel method called Self-Distillation DARTS (SD-DARTS), which is illustrated in Fig. \ref{sd-darts}. SD-DARTS leverages knowledge distillation to guide the training of the supernet. Specifically, it transfers the knowledge learned by the supernet in the previous time step to guide the training of the current supernet. By enforcing DARTS to consider information from previous epochs during optimization, SD-DARTS  reduces the sharpness of the supernet's loss landscape. In contrast to existing approaches, our method utilizes the information from previous epochs instead of generating weight perturbations. Additionally, we allow the supernet to be trained until convergence. These distinguishing factors highlight the novelty and value of our method in improving the performance of DARTS.


Additionally, we recognize that the information provided by a single teacher in SD-DARTS may not be sufficient. Ensemble learning suggests that the predictions of multiple models combined together are often more accurate than those of a single model, as the collective knowledge from multiple models is typically more comprehensive. Inspired by ensemble learning, we introduce the concept of ``voting teachers" to guide the training of the supernet. In practice, we aim to leverage the knowledge from multiple supernets in the previous $K$ time steps and aggregate their predictions through voting. By combining the predictions of these supernets, we generate the final teacher output probability. This integration of SD-DARTS with voting teachers enables us to explore and discover high-quality architectures more effectively.


We conduct a series of experiments to evaluate the effectiveness of our proposed method in improving the performance of DARTS. We utilize our method to search for high-quality architectures and assess their performance on various datasets. On the CIFAR-10 dataset, our architecture achieved a remarkable test error rate of 2.58\%, surpassing the performance of DARTS and lots of its variants. The result highlights the superiority of our approach in producing architectures with improved performance on this dataset. Furthermore, we evaluated the performance of our architecture on the challenging ImageNet dataset and achieved a test error rate of 25.0\%. These results reinforce the effectiveness of our novel self-distillation-based NAS method and demonstrate its advantages over state-of-the-art alternatives. Through these experiments on real datasets, we provide empirical evidence to support the efficacy and superiority of our proposed method in generating high-quality architectures with improved performance.

Our contributions can be summarized as follows:
\begin{itemize}
	\item We propose SD-DARTS, a new DARTS method with better performance. It distills the knowledge from the previous steps of the supernet to guide the training of the supernet in the current step.  It can flatten the supernet's loss landscape to bridge the performance gap between the supernet and the optimal subnet.
	\item We propose voting teachers, a new method by using multiple  teachers to guide the training of the supernet.  The predictions of voting teachers are more accurate than a single teacher. We vote the predications of supernets in the previous $K$ time steps to generate the final teacher output probability and use the teacher output probability to guide the training of the supernet itself.
	\item Extensive experiments demonstrate that our method can effectively improve the performance of DARTS and achieves good results on mainstream datasets.
\end{itemize}

\section{Related Work}

\subsection{Differentiable Architecture Search}

\textit{Differentiable Architecture Search} (DARTS) has gained popularity as a low-cost NAS method by formulating NAS as a continuous optimization problem. Several methods have been proposed to enhance the performance of DARTS. P-DARTS \cite{DBLP:journals/ijcv/ChenXWT21} gradually increases the depth of the supernet to address the discrepancy between architecture depths in search and evaluation scenarios. R-DARTS \cite{zela2019understanding} employes early stopping based on hessian eigenvalues to monitor changes in the optimization process. SDARTS \cite{chen2020stabilizing} introduces perturbations on architecture parameters to control hessian eigenvalues and flatten the supernet's loss landscape. DARTS+ \cite{https://doi.org/10.48550/arxiv.1909.06035} incorporates early stopping when the number of skip connections in the normal cell exceeds a threshold. DARTS-PT \cite{wang2021rethinking} introduces an alternative perturbation-based architecture selection method. DAAS \cite{tian2021discretization} introduces an entropy-based loss term to guide the super-network towards the desired topology. Gold-nas \cite{bi2020gold} employes a variable resource constraint in one-level optimization to gradually prune out weaker operators. DARTS- \cite{chu2020darts} addes an auxiliary skip connection to ensure fair competition among all operations.  Different from these works, we aim to alleviate the discretization gap of DARTS by focusing on flattening the supernet's loss landscape with  less computational cost.

\subsection{Self Distillation}

Knowledge Distillation (KD) \cite{hinton2015distilling} is a technique that transfers the knowledge of a well-trained teacher model to a student model during training. Traditionally, a well-trained teacher model with higher capacity than the student model is used to guide the training process. However, obtaining a well-trained teacher model can be challenging. To address this issue, recent works \cite{zhang2019your, yun2020regularizing} have explored the use of self-distillation, where the student model itself is used as a teacher to distill its own knowledge.

Self-distillation has gained popularity in various domains. For instance, Mean Teacher \cite{tarvainen2017mean} combines self-distillation with semi-supervised learning to improve performance. It minimizes the L2 distance between the teacher model's predictions and the student model's predictions during semi-supervised learning, where the teacher model is the student model itself.  Dual-Teacher++ \cite{li2020dual} introduces a state-of-the-art semi-supervised domain adaptation framework using dual teacher models. It includes an inter-domain teacher model that explores cross-modality priors from the source domain and an intra-domain teacher model that investigates the knowledge within the unlabeled target domain. LE-UDA \cite{zhao2022uda} introduces a self-ensembling consistency approach for knowledge transfer in unsupervised domain adaptation, combining it with a self-ensembling adversarial learning module to achieve better feature alignment. Li et al. \cite{li2021hierarchical} proposes a hierarchical consistency regularized mean teacher framework for 3D left atrium segmentation, where the student model is optimized using multi-scale deep supervision and hierarchical consistency regularization. ACT-NET \cite{zhao2022act} advances teacher-student learning with a co-teacher network, facilitating knowledge distillation from large models to small ones by alternating student and teacher roles. ODC \cite{wei2019online} applies self-distillation to neural machine translation tasks, updating the teacher model when its performance surpasses the current teacher model on the validation data. In our paper, we aim to combine self-distillation with DARTS to enhance the performance of architecture search. By leveraging the knowledge distilled from the student model itself, we expect to improve the quality of the generated architectures in DARTS.


\subsection{Sharpness Aware Minimisation}


Sharpness-aware minimization (SAM) \cite{foret2021sharpnessaware} is a regularization technique that aims to flatten a network's loss landscape. SAM achieves this by introducing adversarial noise to the network parameters during training, which encourages the loss to be more stable and less sensitive to small parameter perturbations. By promoting a flatter loss landscape, SAM can enhance the generalization ability of the model. ASAM \cite{pmlr-v139-kwon21b} builds upon SAM by utilizing a generalization bound to establish a connection between the sharpness of the loss landscape and the generalization gap. By minimizing the sharpness of the loss landscape, ASAM aims to reduce overfitting and improve the model's ability to generalize to unseen data. Zhao et al. \cite{2022arXiv220203599Z} proposes to penalize the gradient norm of the loss function during optimization to flatten the network's loss landscape. By regularizing the gradient norm, the optimization process becomes more stable and less prone to sharp fluctuations, leading to a flatter loss landscape. SAF \cite{du2022sharpness} addresses the issue of sudden drops in loss that can occur in sharp local minima during the trajectory of weight updates. SAF aims to avoid these sudden drops by dynamically adjusting the learning rate based on the sharpness of the loss landscape. By preventing sharp drops in loss, SAF promotes a flatter optimization process and helps the model converge to better solutions. These approaches highlight the importance of considering the flatness of the loss landscape in training deep neural networks. By promoting a flatter and more stable loss landscape, these techniques aim to improve the generalization ability and convergence properties of the models.

\section{Methodology}

\subsection{Preliminary}

Differentiable Architecture Search (DARTS) \cite{liu2018darts} is a popular Neural Architecture Search (NAS) method known for its high search efficiency and low computational cost. Unlike searching for the entire network, DARTS focuses on searching for cells within the network. Each cell is represented as a directed acyclic graph (DAG) with multiple nodes, where each node represents a feature map. The cell has two input nodes, one output node, and several intermediate nodes. The information from the input nodes is propagated through the intermediate nodes to the output node using operations associated with directed edges. To determine the importance of different operations in the candidate operation space $\mathcal{O}$, DARTS assigns weights to each operation using architecture parameters $\alpha^{(i,j)}$. The weighted sum of operations is computed as $\bar{o}^{(i,j)}(x) = \sum_{o \in \mathcal{O}} \frac{\exp \left(\alpha_{o}^{(i, j)}\right)}{\sum_{o^{\prime} \in \mathcal{O}} \exp \left(\alpha_{o^{\prime}}^{(i, j)}\right)} o(x)$. The architecture parameters control the importance of each operation on the corresponding edge.

The search process of DARTS can be formulated as a bilevel optimization problem, where the goal is to minimize the validation loss $\mathcal{L}_{val}$ with respect to the architecture parameters $\alpha$, while finding the optimal network parameters $w^*(\alpha)$ that minimize the training loss $\mathcal{L}_{train}$ with respect to $w$. This is expressed as:

\begin{equation}
	\begin{aligned}
		&\min_{\alpha} \mathcal{L}_{val}(w^*(\alpha), \alpha) \\
		&\text{s.t. } w^*(\alpha) = \arg\min_{w} \mathcal{L}_{train}(w, \alpha).
	\end{aligned}
	\label{eq_sd-darts}
\end{equation}

After optimizing this bilevel optimization problem, the supernet is discretized to derive the optimal architecture based on the architecture parameters. This is done by selecting the operation with the highest weight for each edge, i.e., $o^{(i,j)} = \arg\max_{o \in \mathcal{O}} \alpha_{o}^{(i, j)}$. The discretization step allows us to obtain the final architecture based on the learned architecture parameters, enabling the deployment of the optimized network architecture.


\subsection{Motivation}

Recent studies \cite{zela2019understanding, chen2020stabilizing} shows that there is a performance gap between the supernet and the optimal subnet, and an important factor that results in discretization gap of Differentiable Architecture Search (DARTS) is the sharpness of the supernet's loss landscape. It has been observed that when the supernet's loss landscape is flatter, DARTS is more likely to discover architectures with better performance through the discretization process. Conversely, when the supernet's loss landscape is sharp, the found architectures have worse performance. However, the existing methods proposed in these papers have limitations in effectively and efficiently addressing the problem of sharpness in the supernet's loss landscape. More research is needed to develop more effective and efficient techniques to alleviate this issue and further improve the performance of DARTS.


Several recent papers \cite{foret2021sharpnessaware, pmlr-v139-kwon21b} have explored the idea of flattening the loss landscape by attaching adversarial noise to network parameters during the training stage. Building on this concept, SDARTS \cite{chen2020stabilizing} introduces the attachment of adversarial noise to architecture parameters during the search stage to enhance the flatness of the supernet's loss landscape. However, this method incurs significant computational cost in generating the adversarial noise. Another approach introduced by SAF \cite{du2022sharpness} suggests that minimizing the difference in loss between two consecutive iterations is equivalent to minimizing the sharpness of the loss landscape. Motivated by the concept of self-distillation, which utilizes the network's own information to guide its training, we employ self-distillation as a framework to guide the training of the supernet. This framework aims to enhance the flatness of the supernet's loss landscape.

\subsection{Self-Distillation DARTS}

In our paper, we propose the self-distillation DARTS (SD-DARTS) method, which utilizes the self-distillation framework to guide the training of the supernet in DARTS. SD-DARTS aims to improve the performance of the supernet by attaching self-distillation on the inner-level problem of DARTS's bi-level optimization. The formulation of SD-DARTS is as follows:
\begin{equation}
	\begin{aligned}
		&\min \limits _{\alpha}   \mathcal{L}_{val} (w^*(\alpha ), \alpha)  + \lambda  \mathcal{H} \left(f^S(x), f^T(x) \right) \\
		&\text { s.t. } w^{*}(\alpha)=\arg\min \limits_{w} \mathcal{L}_{\text {train }} (w, \alpha) + \lambda  \mathcal{H} \left(f^S(x), f^T(x) \right),
		\label{eq_sd_darts}
	\end{aligned}
\end{equation} 
In  Eq. \eqref{eq_sd_darts}, $f$ represents the supernet of DARTS, $f^S$ and $f^T$ are the student and teacher models, respectively. The student model corresponds to the supernet at time step $t$, i.e., $f^S = f_t(x)$, and the teacher model corresponds to the student model at the previous time step. The output probabilities of the student and teacher models are denoted as $f^S(x)$ and $f^T(x)$, respectively. The metric $\mathcal{H}$ is used to calculate the correlation between the output probabilities, and in our paper, we use the Kullback-Leibler (KL) divergence as the metric. The regularization coefficient $\lambda$ is used to balance the importance of the two loss terms. In SD-DARTS, the teacher at the present time step is the supernet at the previous time step, i.e.,
\begin{equation*}
	f^T(x) = f_{t-1}(x).
\end{equation*} 
Thus, the formulation in Equation (\ref{eq_sd_darts}) can be simplified as:
\begin{equation*}
	\begin{aligned}
		&\min \limits _{\alpha}   \mathcal{L}_{val}(w^*(\alpha ), \alpha)  + \lambda  \mathcal{H} \left(f_t(x), f_{t-1}(x)\right) \\
		&\text { s.t. } w^{*}(\alpha)=\arg \min\limits_{w} \mathcal{L}_{\text {train }}(w, \alpha) + \lambda  \mathcal{H} \left(f_t(x), f_{t-1}(x)\right).
	\end{aligned}
\end{equation*} 
The detailed process of SD-DARTS is illustrated in Fig. \ref{sd-darts}. Here we will introduce  SD-DARTS in detail.


Firstly, we start by warming up the supernet $f$. Both the architecture parameters $\alpha$ and the network parameters $w$ are randomly initialized. The goal of this warm-up phase is to allow the supernet to learn valuable knowledge from the training data. It is important to note that only high-quality supernets can serve as qualified teachers to guide the training of the supernet itself. This is because low-quality teachers may provide misleading guidance and hinder the training process. Therefore, the warm-up phase plays a crucial role in preparing the supernet for effective self-distillation.


After the warm-up phase, we introduce a self-distillation framework to guide the training of DARTS. At epoch $t$, we utilize the supernet at epoch $t-1$ as the teacher model, denoted as $f^T$, where the previous supernet itself becomes the teacher. To obtain the teacher output probability, there are two approaches.  The first approach involves storing the parameters of the teacher model in the GPU memory and generating the teacher output probability by running the teacher model when needed. However, this approach requires significant computational resources. To mitigate the computational cost, we propose an alternative approach. We store the output probability of the supernet at the previous time step, i.e., $f^T(x)$, directly in the memory and utilize it when needed. In practice, computer memory capacity is typically large, allowing for the allocation of a small portion of memory to store the student output probability. This approach effectively reduces the computational consumption associated with accessing the teacher output probability.



Next, we calculate the correlation between the teacher output probability, $f^T(x)$, and the student output probability, $f_t(x)$, at epoch $t$, using the chosen correlation metric $\mathcal{H}$. There are various correlation metrics available, such as Euclidean Distance (ED), Manhattan Distance (MD), Cosine Distance (CD), and Kullback-Leibler divergence (KL). Since our task is image classification, we utilize KL divergence as the correlation metric $\mathcal{H}$ in our paper. KL divergence is a simple metric used to quantify the differences between two probability distributions. In our case, we calculate the KL divergence as follows:

\begin{equation*}
	\mathcal{H} (f^S(x), f^T(x)) = \sum_i f^S(x_i) \log \frac{f^S(x_i)}{f^T(x_i)},
\end{equation*}
where $f(x_i)$ denotes the output probability for the $i$-th example. We use this correlation metric as a regularization term and incorporate it into the training of the supernet. 

\begin{algorithm*}[t]
	\caption{SD-DARTS}
	\label{alg:sd-darts}
	\begin{algorithmic}
		\STATE {\bfseries Input:} supernet $f$, correlation metric $\mathcal{H}$, total epochs $E$, warm-up epochs $\xi$.
		\STATE Initialize architecture parameters $\alpha$ and network parameters $w$.
		\STATE Warm up the supernet $f$ for $\xi$ epochs.
		\FOR{t $\gets$ ($\xi+1$) to $E$ }
		\STATE Select the  supernet $f_{t-1}$ at the previous time step (epoch $t-1$) as the teacher. 
		\STATE Update $\alpha$ by optimizing $\min \limits _{\alpha} \mathcal{L}_{val}(w^*(\alpha ), \alpha)  + \lambda  \mathcal{H} \left(f_t(x), f_{t-1}(x)\right) $;
		\STATE Update $w$ by optimizing $\min \limits_{w} \mathcal{L}_{\text {train }}(w, \alpha) + \lambda  \mathcal{H} \left(f_t(x), f_{t-1}(x)\right) $;
		\ENDFOR
	\end{algorithmic}
\end{algorithm*}

The overall process of SD-DARTS is summarized in Algorithm \ref{alg:sd-darts}. Similar to DARTS, our method also uses alternative optimization strategy to optimize architecture parameters and network parameters. SGD and Adam are used to train network parameters and architecture parameters, respectively. The experiment settings in detail are shown in Sec. \ref{as}. After the training of the supernet, the optimal architecture is derived from the supernet based on architecture parameters.

  Here, we demonstrate how self-distillation can effectively flatten the loss landscape of an universal neural network $f$ with its parameters denoted as $\alpha$ by employing theoretical tools from SAM \cite{foret2021sharpnessaware}  and SAF \cite{du2022sharpness}.   SAM \cite{foret2021sharpnessaware} is introduced to flatten the loss landscape by solving the following minmax problem:
\begin{equation*}
	\min _{\alpha} \max _{\|\epsilon\|_2 \le \rho} \mathcal{L}(f_{\alpha + \epsilon}).
\end{equation*}
where $\mathcal{L}$, $\rho$ and $\epsilon$ denote the loss function,  allowed perturbation constraint and  perturbation on  parameters, respectively. Further, the sharpness loss $R(f_{\alpha})$ can be represented as $\mathcal{L}(f_{\alpha+\hat{\epsilon}}) - \mathcal{L}(f_{\alpha}) $,  where $\hat{\epsilon}$  denotes the optimal perturbation on  parameters, and $\hat{\epsilon}$ is calculated by $\rho \frac{\nabla_\alpha \mathcal{L}(f_{\alpha})^\top}{\| \nabla_\alpha \mathcal{L}(f_{\alpha}) \|_2} $ \cite{foret2021sharpnessaware}. We use a first-order Taylor expansion to decompose the sharpness loss as follows:
\begin{align}
	R(f_{\alpha})& = \mathcal{L}(f_{\alpha+\hat{\epsilon}}) - \mathcal{L}(f_{\alpha}) 
	\approx \mathcal{L}(f_{\alpha}) +\hat\epsilon  \nabla_\alpha \mathcal{L}(f_{\alpha}) - \mathcal{L}(f_{\alpha}) \nonumber\\
	&= \rho \frac{\nabla_\alpha \mathcal{L}(f_{\alpha})^\top}{\| \nabla_\alpha \mathcal{L}(f_{\alpha}) \|_2}  \nabla_\alpha \mathcal{L}(f_{\alpha})
	= \rho \| \nabla_\alpha L(f_{\alpha}) \|_2 \label{sharp_grid},
\end{align}
 Eq. \eqref{sharp_grid} shows that minimizing the  sharpness loss  is equivalent to minimizing the  $\ell_2$-norm of the gradient $\nabla_\alpha L(f_{\alpha})$. 

\begin{figure*}[t]
	\vskip 0.2in
	\begin{center}
		\centerline{\includegraphics[width=10cm]{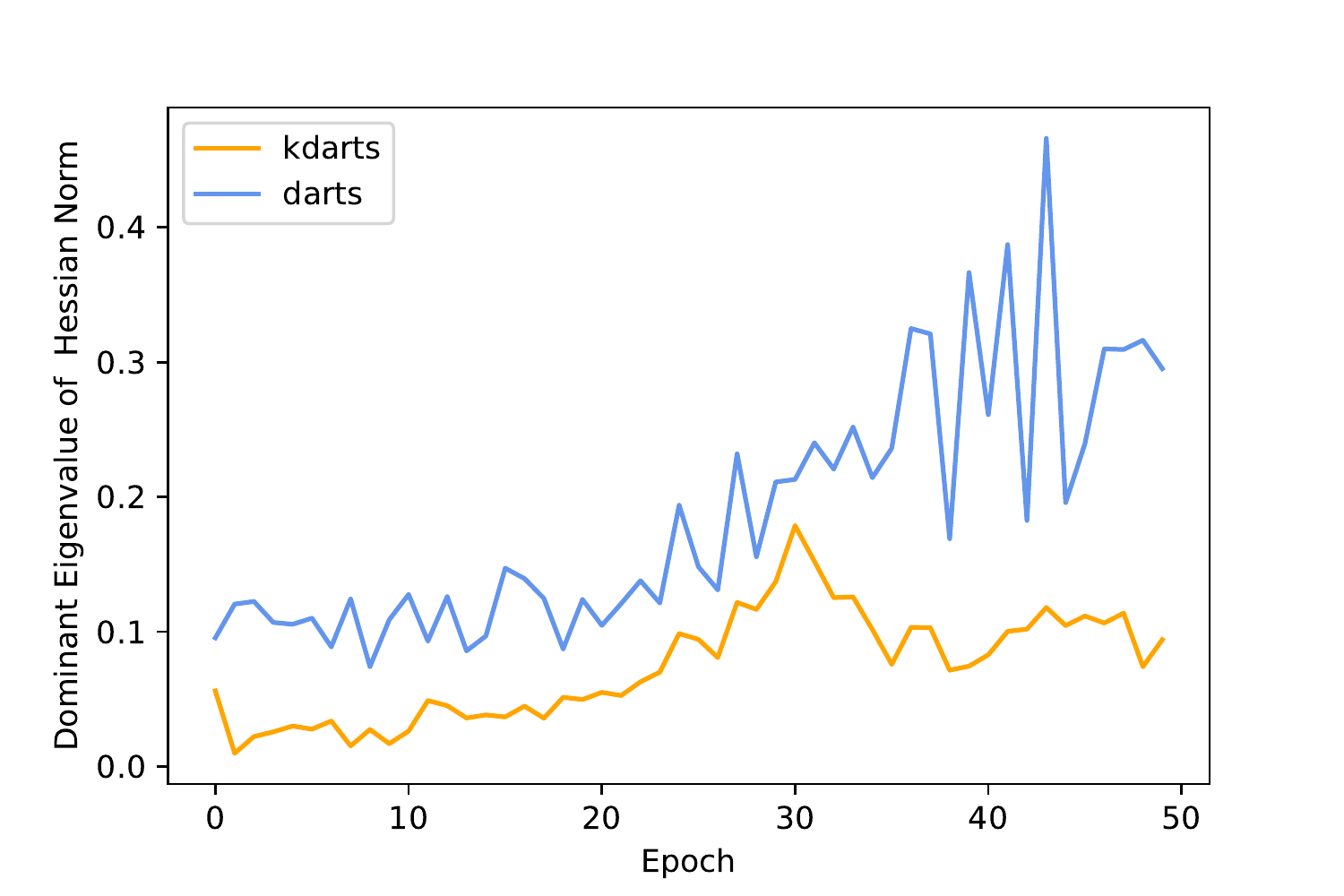}}
		\caption{Trajectory  of the supernet's Hessian norm on standard DARTS and SD-DARTS.  Dominant Eigenvalue of the supernet's Hessian norm on SD-DARTS is smaller than standard DARTS, and the result shows that the loss landscape of SD-DARTS is  flatter than DARTS. }
		\label{ev}
	\end{center}
	\vskip -0.2in
\end{figure*}

On the other hand,  motivated by SAF \cite{du2022sharpness}, the change of the network's validation loss in the two consecutive iterations can be represented as
\begin{equation}
	\label{sharp_grid2}
	\begin{aligned}
		& \mathcal{L}(f_{\alpha}) - \mathcal{L}(f_{\alpha - \lambda \nabla_\alpha \mathcal{L}(f_{\alpha})}) \approx \lambda \| \nabla_\alpha \mathcal{L}(f_{\alpha}) \|^2_2 \approx \frac{\lambda}{\rho^2} R(f_{\alpha})^2 . \\
	\end{aligned} 
\end{equation}
where $\lambda$ denotes as the learning rate in the training of parameters. SAM \cite{foret2021sharpnessaware} shows that the learning rate $\lambda$ in the training of parameters is typically smaller than $\rho$. Based on Eq. \eqref{sharp_grid2}, we find that the change of the loss  is proportional to $R(f_{\alpha})^2$. Hence, minimizing the loss difference is equal to minimizing the  sharpness  of the network in this case. Our analysis can be easily   generalized to DARTS. Fig. \ref{ev} also shows that dominant eigenvalue of the supernet's Hessian norm on SD-DARTS is smaller than standard DARTS, i.e.,  the loss landscape of SD-DARTS is  flatter than DARTS.

\begin{remark}
	In our method, we utilize self-distillation (SD) as a framework to train the supernet. To assess the sharpness of the validation loss landscape, we compute the dominant eigenvalue of the supernet's Hessian norm. Fig. \ref{ev} demonstrates that the dominant eigenvalue of the supernet's Hessian norm in SD-DARTS is smaller than that in standard DARTS. This indicates that the loss landscape of SD-DARTS is flatter compared to DARTS. The flatter loss landscape in SD-DARTS suggests that our method is effective in reducing the sharpness and improving the stability of the supernet's training process.
\end{remark}

\begin{figure*}[t]
	\vskip 0.2in
	\begin{center}
		\centerline{\includegraphics[width=\linewidth]{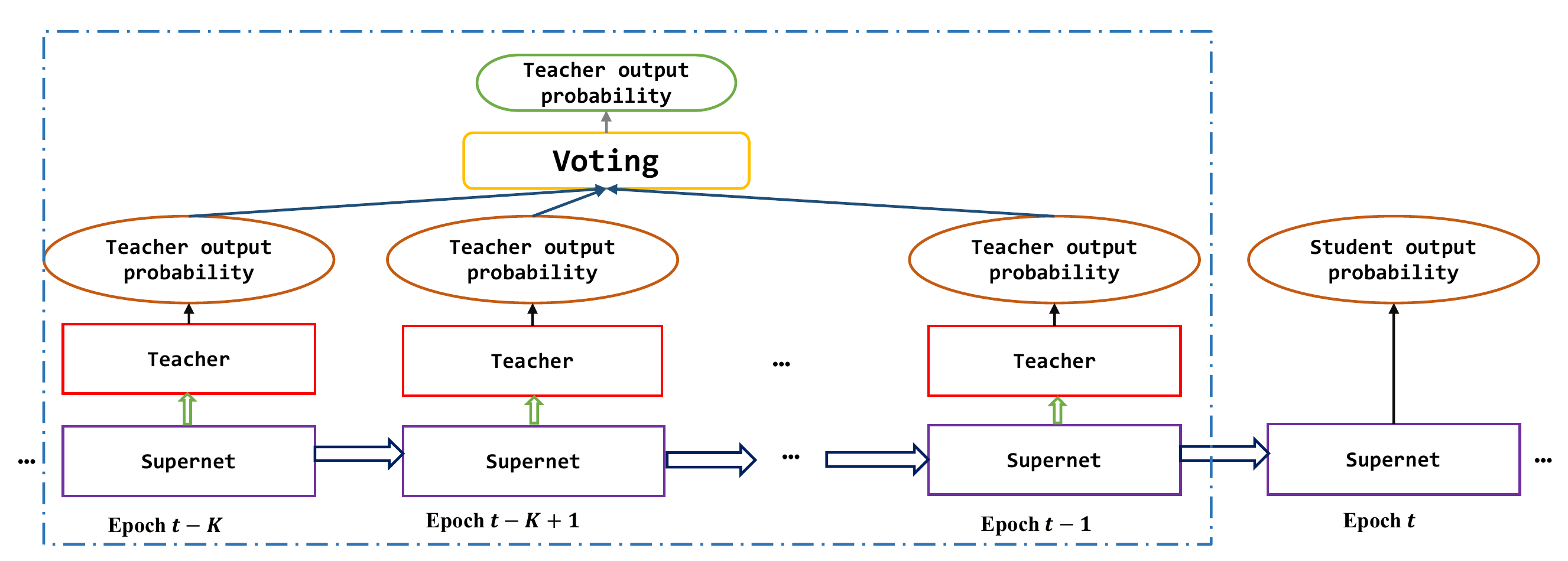}}
		\caption{Voting teachers.  We utilize supernets at previous $K$ time steps as teachers and then vote teacher output probabilities outputted by  supernet to generate the final teacher output probability.}
		\label{voting}
	\end{center}
	\vskip -0.2in
\end{figure*}

\subsection{Voting Teachers}


In SD-DARTS, we utilize the supernet from the previous time step as the teacher to guide the training process. However, we recognize that the information provided by a single teacher may be limited. To overcome this limitation, we propose to utilize multiple teachers in order to obtain a more diverse and rich source of information.  Each teacher contributes its own expertise and experience to guide the training of the supernet. To make predictions, we employ a voting scheme where the output probabilities of each teacher are combined to generate the final teacher prediction. This approach of utilizing multiple teachers allows us to leverage a broader range of knowledge and increase the diversity of guidance during the training process. By aggregating the insights from multiple teachers, we can potentially achieve better performance and more robust architectures in SD-DARTS.


\begin{algorithm*}[t]
	\caption{Voting Teachers}
	\label{alg:voted-teacher}
	\begin{algorithmic}
		\STATE {\bfseries Input:} supernet $f$, correlation metric $\mathcal{H}$, total epochs $E$, warm-up epochs $\xi$, time window $K$.
		\STATE Initialize architecture parameters $\alpha$ and network parameters $w$.
		\STATE Warm up the supernet $f$ for $\xi$ epochs.
		\FOR{t $\gets$ ($\xi+1$) to $E$ }
		\STATE Select the  supernets  at the previous $K$ time steps (from epoch $t-K$ to epoch $t-1$) as the teachers. 
		\STATE Update $\alpha$ by optimizing $\min \limits _{\alpha} \mathcal{L}_{val}(w^*(\alpha ), \alpha)  + \lambda  \mathcal{H} \left(f(x), \frac{1}{K} \sum \limits_{i=1}^{K}  f_{t-i}(x)\right) $;
		\STATE Update $w$ by optimizing $\min \limits_{w} \mathcal{L}_{\text {train }}(w, \alpha) + \lambda  \mathcal{H} \left(f(x), \frac{1}{K} \sum \limits_{i=1}^{K}  f_{t-i}(x)\right) $;
		\ENDFOR
	\end{algorithmic}
\end{algorithm*}

In our proposed method, known as SD-DARTS, we introduce a concept called ``voting teachers" to guide the training of the supernet. Inspired by ensemble learning, which demonstrates that the performance of multiple models is often better than that of a single model, we leverage the idea of combining multiple teachers to enhance the training process. In the context of SD-DARTS, we consider a time window of size $K$, where we select multiple supernets from the previous $K$ time steps as teachers. The output probabilities of these teacher supernets are then averaged to generate the final teacher output probability. Mathematically, the final teacher output probability can be represented as follows:
	\begin{equation*}
		f^T(x) = \frac{1}{K} \sum \limits_{i=1}^{K}  f_{t-i}(x),
	\end{equation*}
where $f_{t-i}(x)$ denotes the output probability of the supernet at time step $t-i$. By combining the predictions from multiple teachers, we aim to benefit from their collective knowledge and insights, leading to improved guidance in the training process of the supernet. This approach allows us to leverage the strengths of different teacher supernets and potentially achieve better performance and more robust architectures in SD-DARTS.


In our proposed method, which combines voting teachers with SD-DARTS, we aim to leverage the knowledge and insights from multiple teacher supernets to guide the search for architectures. The objective function in this case is formulated as follows:
	\begin{equation*}
		\begin{aligned}
			&\min \limits _{\alpha}   \mathcal{L}_{val}(w^*(\alpha ), \alpha)  + \lambda  \mathcal{H} \left(f(x), \frac{1}{K} \sum \limits_{i=1}^{K}  f_{t-i}(x)\right) \\
			&\resizebox{.9\hsize}{!}{$\text { s.t. } w^{*}(\alpha)=\arg \min\limits_{w} \mathcal{L}_{\text {train }}(w, \alpha) + \lambda  \mathcal{H} \left(f(x), \frac{1}{K} \sum \limits_{i=1}^{K}  f_{t-i}(x)\right),$}
		\end{aligned}
	\end{equation*} 
where $t$ represents the present time step, $\lambda$ represents the regularization coefficient,  $K$ is the time window that controls the number of teachers, and $\frac{1}{K} \sum \limits_{i=1}^{K} f_{t-i}(x)$ denotes the final teacher output probability obtained by averaging the output probabilities of the multiple teacher supernets. The regularization term $ \mathcal{H} \left(f(x), \frac{1}{K} \sum \limits_{i=1}^{K} f_{t-i}(x)\right)$ encourages consistency between the supernet's output probability and the averaged output probabilities of the voting teachers. To implement the voting teachers approach, we provide an algorithmic description in Algorithm \ref{alg:voted-teacher}. This algorithm outlines the steps for selecting the teacher supernets from the previous time steps, averaging their output probabilities, and using the averaged output probability as the final teacher output probability to guide the training of the supernet. By incorporating voting teachers into SD-DARTS, we aim to take advantage of the collective knowledge and insights from multiple teacher supernets, potentially leading to improved performance and more robust architectures during the search process.


By utilizing voting teachers in combination with SD-DARTS, we can leverage the collective knowledge of multiple teacher supernets to guide the search for architectures. This approach allows for more diverse and informed guidance during the training of the supernet, potentially leading to better performance and improved architectures. The additional memory required to store the output probabilities of the teacher supernets is relatively small compared to the overall memory requirements of the training process. This allows us to leverage the rich information from multiple teachers without significantly increasing the computational or memory overhead.

\begin{remark}
	
\begin{enumerate}
		\item Ensemble learning has shown that combining the knowledge of multiple models can lead to improved performance compared to a single model. By selecting multiple supernets from the previous time steps as teachers, we can leverage the diverse knowledge and expertise of these models to guide the training of the supernet. The use of multiple teachers helps to capture a broader range of information and insights, leading to a more comprehensive and robust training process. It allows the supernet to benefit from the collective wisdom of multiple architectures, increasing the chances of finding high-quality architectures.
		\item Our method differs from Mean Teacher \cite{tarvainen2017mean} in the approach of averaging. While Mean Teacher focuses on averaging the weights of the network, our method aims to average the output of the network. We have two reasons for choosing the voting teachers approach over Mean Teacher. 	Firstly, the parameter scale of the network's output is significantly smaller than the weights, and computational and memory overhead will become bigger when we average weights instead of outputs. Because weights and outputs are very closely linked, by using the average of the output to guide our self-distillation training, our method still has an impact on the weights. It allows us to leverage the benefits of the output averaging while considering the relative importance of the weights. Secondly, the coupling between the architecture parameters and network parameters in our model, due to alternative optimization strategies, can lead to instability during the training process when using teachers generated by averaging the weights. The instability can hinder convergence and affect the overall performance of the model. By considering these factors, we have determined that the voting teachers approach is better suited to our specific objectives and model characteristics. The decision allows us to effectively leverage the advantages of averaging while mitigating potential instabilities in the training process.   
\end{enumerate}
\end{remark}

\section{Experiment}

In the sections, we carry out extensive experiments to verify the effectiveness  of our method. Our experiments are conducted on $3\times$ NVIDIA GeForce GTX 3090 GPUs.  CIFAR-10 \cite{krizhevsky2009learning} and ImageNet \cite{deng2009imagenet} are two image classification datasets which our experiments are conducted on. We search for the optimal architecture by SD-DARTS on CIFAR-10 and then evaluate the architecture on CIFAR-10 and ImageNet.

\begin{figure*}[b]
	\centering
	\begin{tabular}{@{}c@{\hspace{1mm}}c@{\hspace{1mm}}c@{}}
		
		\includegraphics[trim=10mm 13mm 0mm 13mm, clip, width=7cm,height=2.5cm]{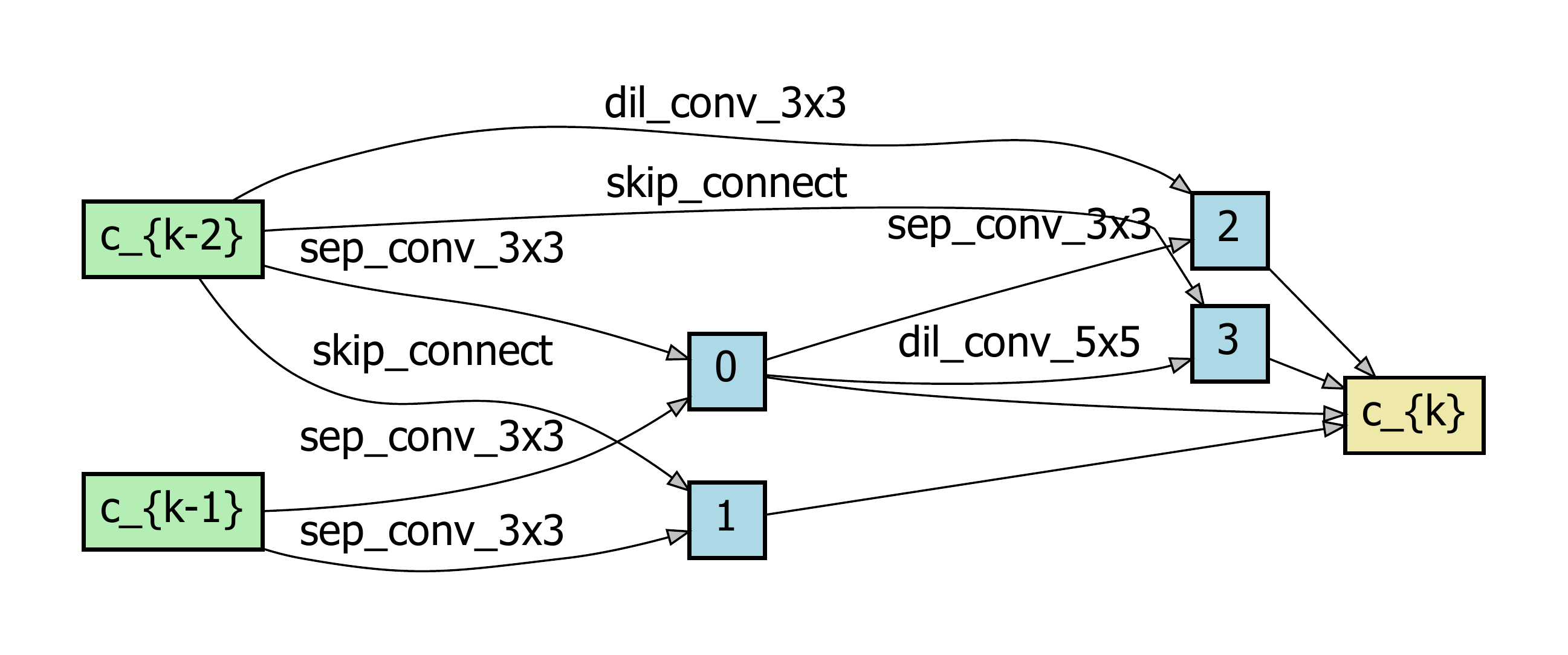} &
		\includegraphics[trim=0mm 13mm 10mm 13mm, clip,  width=7cm,height=2.5cm]{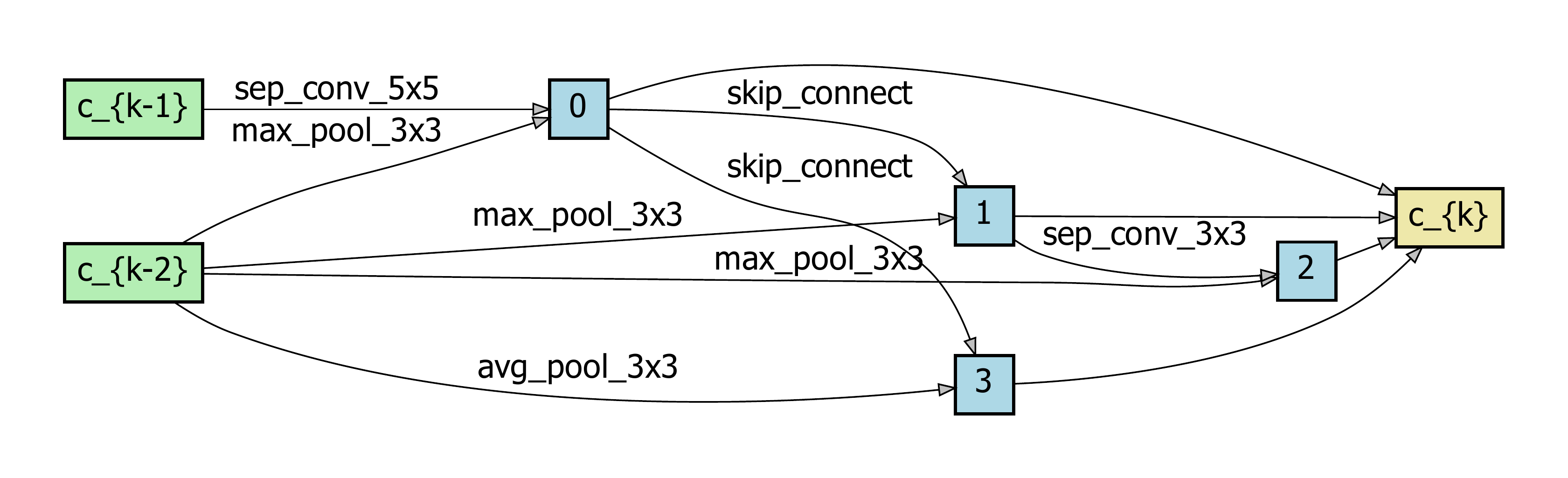} \\
		
		\small (a) Normal cell &
		\small (b) Reduction cell \\
		
	\end{tabular}
	\caption{Cells are searched by SD-DARTS when epochs for warm-up $\xi$ are 25 and  time window $K$ is 2.}
	\label{cell-sddarts}
	\vspace{-10pt}
\end{figure*}

\subsection{Architecture Search}

\label{as}
CIFAR-10 is the dataset that we search for the optimal architecture by SD-DARTS. There are 50K training images and 10K test images in CIFAR-10, and each image's spatial resolution is $32 \times 32$. These images are divided into 10 classes. During the search stage of SD-DARTS, training images are split into two subsets, i.e., the valid subset for optimizing architecture parameters and the training subset for optimizing network parameters. After SD-DARTS searches for the optimal architecture, training images in CIFAR-10 are used to train the optimal architecture and then evaluate the architecture's performance on test images.

The search space of our method is the same as DARTS, and the search space includes 8 candidate operations, $skip\_connect$, $max\_pool\_3\times3$, $avg\_pool\_3\times3$, $sep\_conv\_3\times3$, $sep\_conv\_5\times5$, $dil\_conv\_3\times3$, $dil\_conv\_5\times5$, $zero$. The supernet is stacked by 6 normal cells and 2 reduction cells, which the reduction cells are inserted in $\frac{1}{3}$ and $\frac{2}{3}$ of the total depth of the supernet, respectively. The stride of convolution in the reduction cell is 2; thus, the reduction cell can reduce the spatial resolution of feature maps. However, the stride of convolution in the normal cell is 1, i.e., the spatial size of the feature map is not reduced. Our method aims to search for one normal cell and one reduction cell to build the optimal architecture.

\begin{table}[ht]
	\centering
	\scriptsize
	\caption{Hyperparameter Settings of SD-DARTS.}
	\label{hs_SD-DARTS}
	\begin{tabular}{c|c|c}
		\hline
		\textbf{Hyperparameter} & \textbf{Value} & \textbf{Meaning} \\ \hline
		\textbf{$\xi$}    & 25  & epochs for warm-up   \\ \hline
		\textbf{$\lambda  $}    & 1.0   &  regularization coefficient  \\ \hline
		\textbf{$K$}    & 2 & time window   \\ \hline
		\textbf{$E$}    & 50 & total epochs for search   \\ \hline
		\textbf{$m$}    & 64 & batch size for search   \\ \hline
	\end{tabular}
\end{table}

The train setting of our method is also the same as DARTS. The total epochs $E$ of training the supernet are 50, and the batch size of the train is 64. The initial channel size of the supernet is 16. SGD is used to optimize network parameters $w$with an  initial learning rate  0.025, momentum  0.9 and weight decay  $3\times10^{-4}$. Adam is used to optimize architecture parameters with an initial learning rate  $3\times10^{-4}$, momentum  (0.5, 0.999) and weight decay  $10^{-3}$. We warm up the supernet for 25 epochs, i.e., the epoch for warm-up $\xi$ is 25. Then,  time window $K$ is set as 2, and it means that we will select supernets at the two previous time steps as teachers to guide the training of the supernet. Furthermore, we empirically set the regularization coefficient $\lambda$ as $1.0$ to balance classification loss and correlation loss. Table \ref{hs_SD-DARTS} shows these hyperparameter settings of our method.


\subsection{Architecture Evaluation on CIFAR-10}

Fig. \ref{cell-sddarts} shows cells searched by SD-DARTS, and we stack these cells to build the optimal architecture. In the subsection, we evaluate the performance of the optimal architecture on CIFAR-10. The training setting of our method is also the same as DARTS. We stack 20 cells to build a network with an initial channel size of 36, where two reduction cells are put into $\frac{1}{3}$ and $\frac{2}{3}$ of the total depth of the network. We train the network for 600 epochs with batch size 96. SGD is utilized to optimize the network with an
initial learning rate of 0.025. Furthermore, we also use data augmentation to regularize the network's training, including cutout and auxiliary towers; the cutout length is 16, the weight of auxiliary towers is 0.4, and the probability of path dropout is 0.3. 

\begin{table*}[t]
	\centering
	\scriptsize
	\caption{Comparison with state-of-the-art image classifiers on CIFAR-10.}
	\label{acc_c10}
	
	\begin{tabular}{@{}ccccc@{}}
		\toprule
		\textbf{Architecture} &
		\begin{tabular}[c]{@{}c@{}} \textbf{Test Err.}\\ (\textbf{\%})\end{tabular} &
		\begin{tabular}[c]{@{}c@{}}\textbf{Params}\\ \textbf{(M)}\end{tabular} &
		\multicolumn{1}{l}{\begin{tabular}[c]{@{}l@{}}\textbf{Search Cost}\\ \textbf{(GPU-days)}\end{tabular}} &
		\begin{tabular}[c]{@{}c@{}}\textbf{Search} \\ \textbf{Method}\end{tabular} \\ \midrule
		DenseNet-BC \cite{huang2017densely}    & 3.46  & 25.6 & -    & manual    \\ \midrule
		NASNet-A \cite{zoph2018learning} & 2.65     & 3.3  & 1800 & RL        \\
		AmoebaNet-A \cite{real2019regularized}    & 3.34$\pm$0.06 & 3.2  & 3150 & evolution \\
		AmoebaNet-B \cite{real2019regularized}    & 2.55$\pm$0.05 & 2.8  & 3150 & evolution \\
		PNAS  \cite{liu2018progressive}        & 3.41$\pm$0.09 & 3.2  & 225  & SMBO      \\
		ENAS \cite{pham2018efficient}          & 2.89          & 4.6  & 0.5  & RL        \\ \midrule
		DARTS ($1^{\text{st}}$ order)  \cite{liu2018darts} & 3.00$\pm$0.14 & 3.3  & 0.4  & gradient  \\
		DARTS ($2^{\text{nd}}$ order) \cite{liu2018darts} & 2.76$\pm$0.09 & 3.3  & 1    & gradient  \\
		SNAS (mild)  \cite{xie2018snas}      & 2.98          & 2.9  & 1.5  & gradient  \\
		ProxylessNAS  \cite{cai2018proxylessnas}     & 2.08          & -    & 4    & gradient  \\
		P-DARTS  \cite{DBLP:journals/ijcv/ChenXWT21}       & 2.5           & 3.4  & 0.3  & gradient  \\
		PC-DARTS  \cite{xu2019pc}     & 2.57$\pm$0.07 & 3.6  & 0.1  & gradient  \\
		$\beta$-DARTS \cite{2022arXiv220301665Y} & 2.53$\pm$0.08 & 3.8 & 0.4 & gradient\\ 
		SDARTS-RS \cite{chen2020stabilizing}     & 2.67$\pm$0.03 & 3.4  & 0.4  & gradient  \\
		SDARTS-ADV \cite{chen2020stabilizing} & 2.61$\pm$0.02 & 3.3 & 1.3 & gradient  \\
		GDAS  \cite{dong2019searching}     & 2.93          & 3.4  & 0.3  & gradient  \\
		R-DARTS (L2) \cite{zela2019understanding}       & 2.95$\pm$0.21 & -    & 1.6  & gradient  \\
		SGAS (Cri 1. avg)  \cite{li2020sgas}    & 2.66$\pm$0.24 & 3.7  & 0.25 & gradient  \\
		DARTS-PT \cite{wang2021rethinking}        & 2.61$\pm$0.08 & 3.0  & 0.8  & gradient  \\ 
		DARTS-\cite{chu2020darts} & 2.59$\pm$0.08 & 3.5 & 0.4 & gradient \\ 
		U-DARTS \cite{huang2023u} & 2.59 $\pm$ 0.06 & 3.3 & 4 & gradient \\
		DARTS-PAP \cite{li2023darts} & 2.51 & 3.9 & 0.4 & gradient \\
		\midrule
		SD-DARTS     & 2.58$\pm$0.11  & 3.3  & 0.37   & gradient  \\ 
		SD-DARTS (best) & 2.44 & 3.3  & 0.37   & gradient \\ \bottomrule 	
	\end{tabular}
\end{table*}

Table \ref{acc_c10} shows the performance of the architecture searched by our method on CIFAR-10. The architecture searched by our method achieves a 2.58\% mean test error on CIFAR-10, and it can achieve a 2.44\% best test error when setting the appropriate random seed to train the architecture. The performance of our method is beyond DARTS  and its variants (e.g., DARTS-PT, SGAS, GDAS) a lot. In other words, our method achieves state-of-the-art (SOTA) performance. The magnitude of our architecture's parameters  is 3.3M, and it is equal to the architecture searched by DARTS and less than other variants.

Furthermore, the search cost of our method is 0.37 GPU days, and it costs a few less GPU days than DARTS.  In a word, with limited resources, our method can find the architecture with SOTA performance.

\begin{table}[h]
	\centering
	\scriptsize
	\caption{Comparison with state-of-the-art classifiers on ImageNet.}
	\label{acc_imagenet}
	\begin{tabular}{@{}ccccccc@{}}
		\toprule
		\multirow{2}{*}{\textbf{Architecture}} &
		\multicolumn{2}{c}{\textbf{Test Err.(\%)}} &
		\multirow{2}{*}{\begin{tabular}[c]{@{}c@{}}\textbf{Params}\\ \textbf{(M)}\end{tabular}} &
		\multirow{2}{*}{\begin{tabular}[c]{@{}c@{}}$\times +$\\ \textbf{(M)}\end{tabular}} &
		\multirow{2}{*}{\begin{tabular}[c]{@{}c@{}}\textbf{Search Cost}\\ \textbf{(GPU-days)}\end{tabular}} &
		\multirow{2}{*}{\begin{tabular}[c]{@{}c@{}}\textbf{Search}\\ \textbf{Method}\end{tabular}} \\ \cline{2-3}
		& \textbf{top-1} & \textbf{top-5} &  &  &  &  \\ \midrule
		Inception-v1 \cite{szegedy2015going}         & 30.2          & 10.1              & 6.6     & 1448 & -    & manual    \\
		MobileNet \cite{howard2017mobilenets}   & 29.4                      & 10.5                      & 4.2     & 569  & -    & manual    \\
		ShuffleNet 2x (v1) \cite{zhang2018shufflenet}   & 26.4                      & 10.2                      & $\sim$5 & 524  & -    & manual    \\
		ShuffleNet 2x (v2)  \cite{ma2018shufflenet}  & 25.1                      & -                         & $\sim$5 & 591  & -    & manual    \\ \midrule
		NASNet-A  \cite{zoph2018learning}            & 26                        & 8.4                       & 5.3     & 564  & 1800 & RL        \\
		NASNet-B   \cite{zoph2018learning}           & 27.2                      & 8.7                       & 5.3     & 488  & 1800 & RL        \\
		NASNet-C   \cite{zoph2018learning}           & 27.5                      & 9                         & 4.9     & 558  & 1800 & RL        \\
		AKD \cite{1999} & 27.9 & 8.3 & - & - & - & RL \\
		AmoebaNet-A  \cite{real2019regularized}         & 25.5                      & 8                         & 5.1     & 555  & 3150 & evolution \\
		AmoebaNet-B    \cite{real2019regularized}       & 26                        & 8.5                       & 5.3     & 555  & 3150 & evolution \\
		AmoebaNet-C    \cite{real2019regularized}       & 24.3                      & 7.6                       & 6.4     & 570  & 3150 & evolution \\
		FairNAS-A  \cite{chu2021fairnas}           & 24.7                      & 7.6                       & 4.6     & 388  & 12   & evolution \\
		PNAS   \cite{liu2018progressive}               & 25.8                      & 8.1                       & 5.1     & 588  & 225  & SMBO      \\
		MnasNet-92 \cite{tan2019mnasnet}           & 25.2                      & 8                         & 4.4     & 388  & -    & RL        \\ \midrule
		DARTS($2^{\text{nd}}$ order) \cite{liu2018darts} & 26.7                      & 8.7                       & 4.7     & 574  & 4.0  & gradient  \\
		SNAS (mild)  \cite{xie2018snas}          & 27.3                      & 9.2                       & 4.3     & 522  & 1.5  & gradient  \\
		ProxylessNAS \cite{cai2018proxylessnas}         & 24.9                      & 7.5                       & 7.1     & 465  & 8.3  & gradient  \\
		P-DARTS \cite{DBLP:journals/ijcv/ChenXWT21}              & 24.4                      & 7.4                       & 4.9     & 557  & 0.3  & gradient  \\
		PC-DARTS  \cite{xu2019pc}             & 25.1                      & 7.8                       & 5.3     & 586  & 0.1  & gradient  \\
		$\beta$-DARTS \cite{2022arXiv220301665Y} & 24.2 & 7.1 & 5.5 & 609 &0.4 & gradient\\
		SGAS (Cri.1 avg.)  \cite{li2020sgas}   & 24.41                     & 7.29                      & 5.3     & 579  & 0.25 & gradient  \\
		GDAS         \cite{dong2019searching}         & 26.0                      & 8.5                       & 5.3     & 581  & 0.21 & gradient  \\
		SDARTS-RS \cite{chen2020stabilizing}         & 25.6                      & 8.2                       & -     & - & - & gradient  \\
		SDARTS-ADV \cite{chen2020stabilizing}         & 25.2                      & 7.8                      & -     & - & - & gradient  \\
		DARTS-PT \cite{wang2021rethinking} & 25.5 & 8& 4.6&-&0.8&gradient\\
		U-DARTS \cite{huang2023u} & 26.1 & 8.1 & 4.9 & 582 &3 & gradient \\
		DARTS-PAP \cite{li2023darts} & 25.41 & 8.04 & 5.4 & 629 & 0.4 & gradient \\
		 \midrule
		SD-DARTS   &   25.0   &     7.62        &  4.7  &  534    & 0.37  & gradient   \\ \bottomrule
	\end{tabular}
\end{table}

\subsection{Architecture Evaluation on ImageNet}

In the subsection, we utilize ILSVRC2012 \cite{russakovsky2015imagenet} to evaluate the transferability of our method. ILSVRC2012 is a lightweight version of ImageNet with 1,000 classes. It contains 1.28M train images and 50K valid images, and each image's spatial resolution is 224$\times$224. 

Fig. \ref{cell-sddarts} shows the cells searched by our method on CIFAR-10. We stack the cells to build a network and evaluate the network's performance on ImageNet. The train setting of our method on ImageNet is the same as DARTS. We utilize 12 normal cells and 2 reduction cells to build the network, and the reduction cells are inserted into $\frac{1}{3}$ and $\frac{2}{3}$  of the total depth of the network. The initial channel size of the network is 48. We train the network for 250 epochs with batch size 1024 on 3$\times$ NVIDIA GeForce  GTX 3090 GPUs. We utilize SGD to optimize the network with an initial learning rate of 0.5,  a momentum of 0.9, and a weight decay of $3\times10^{-5}$. Furthermore, we utilize an auxiliary loss tower to help train our network with an auxiliary weight of 0.4. We spend three GPU days training the network on 3$\times$ NVIDIA GeForce  GTX 3090 GPUs.


Table \ref{acc_imagenet} demonstrates the evaluation results of our network on the ImageNet dataset, where we achieve a test error rate of 25.0\%. This performance indicates that our method's transferability is superior to DARTS and some of its variants. It is worth noting that certain DARTS variants exhibit lower test errors than our method. However, our architecture has fewer parameters compared to most of its variants. While there may be baseline methods with better results, it is important to consider that their goals might differ from ours. For instance, methods like SGAS \cite{li2020sgas} and P-DARTS \cite{DBLP:journals/ijcv/ChenXWT21} address weight sharing and the depth gap between search and evaluation scenarios, respectively. In contrast, our focus is on flattening the supernet's loss landscape. Furthermore, our method outperforms our concurrent works  in terms of both performance and search cost efficiency. These results demonstrate the value and efficiency of our approach.

 While our experimental results on the ImageNet dataset may not be as strong as desired, we believe that they still provide valuable insights into the performance of our method. Prior works such as SDARTS \cite{chen2020stabilizing} and SAM \cite{foret2021sharpnessaware} have also demonstrated that the dominant eigenvalue of the Hessian norm is a reliable indicator for characterizing the sharpness of the loss landscape. The observation that the dominant eigenvalue of the supernet's Hessian norm in our method is smaller than in standard DARTS, as well as the analysis showing that self-distillation can flatten the loss landscape, provide strong evidence that our proposed algorithm effectively reduces the discretization gap.  Combining these pieces of evidence, we can confidently assert that our method successfully mitigates the discretization gap and flattens the supernet's loss landscape.  Prior works such as R-DARTS \cite{zela2019understanding} and SDARTS \cite{chen2020stabilizing} have shown that a high level of sharpness in the supernet's loss landscape can have a detrimental effect on the performance of DARTS. These works have proposed techniques like early stopping and parameter perturbation to address this issue. However, early stopping may limit the exploration of architectures with better performance, and generating parameter perturbations can be time-consuming. In comparison, our method is more efficient as we only need to preserve the logits from previous training steps. This approach allows us to take advantage of the knowledge accumulated in earlier steps without the need for expensive parameter perturbations. By leveraging self-distillation and voting teachers, our method effectively flatten the supernet's loss landscape and achieves better performance in a more efficient manner compared to prior works such as  SDARTS.

\begin{table}[h]
	\begin{center}
		\setlength\tabcolsep{2pt}
		\setlength\tabcolsep{4pt}
		\caption{Performance comparison on NAS-Bench-201 benchmark~\cite{2020arXiv200100326D}. Note that SD-DARTS only searches on CIFAR-10 dataset, but can  achieve new SOTA on CIFAR-10, CIFAR-100 and ImageNet16-120.}
		\label{201}
		\scriptsize
		\begin{tabular}{ccccccc}
			\hline

			\multirow{2}{*}{Methods} &  \multicolumn{2}{c}{CIFAR-10}              & \multicolumn{2}{c}{CIFAR-100}             & \multicolumn{2}{c}{ImageNet16-120}        \\ \cline{2-7} 
			 & valid               & test                & valid               & test                & valid               & test                \\ \hline
			DARTS(1st)~\cite{liu2018darts}                                       & 39.77±0.00          & 54.30±0.00          & 15.03±0.00          & 15.61±0.00          & 16.43±0.00          & 16.32±0.00          \\
			DARTS(2nd)~\cite{liu2018darts}                                      & 39.77±0.00          & 54.30±0.00          & 15.03±0.00          & 15.61±0.00          & 16.43±0.00          & 16.32±0.00          \\
			GDAS~\cite{dong2019searching}                                          & 89.89±0.08          & 93.61±0.09          & 71.34±0.04          & 70.70±0.30          & 41.59±1.33          & 41.71±0.98          \\
			SNAS~\cite{xie2018snas}                                                 & 90.10±1.04          & 92.77±0.83          & 69.69±2.39          & 69.34±1.98          & 42.84±1.79          & 43.16±2.64          \\
			DSNAS~\cite{Hu_2020_CVPR}                                            & 89.66±0.29          & 93.08±0.13          & 30.87±16.40         & 31.01±16.38         & 40.61±0.09          & 41.07±0.09          \\
			PC-DARTS~\cite{xu2019pc}                                            & 89.96±0.15          & 93.41±0.30          & 67.12±0.39          & 67.48±0.89          & 40.83±0.08          & 41.31±0.22          \\
			iDARTS~\cite{pmlr-v139-zhang21s}                                     & 89.86±0.60          & 93.58±0.32          & 70.57±0.24          & 70.83±0.48          & 40.38±0.59          & 40.89±0.68          \\
			DARTS-~\cite{chu2021darts}                                              & 91.03±0.44          & 93.80±0.40          & \textbf{71.36±1.51}          & 71.53±1.51          & 44.87±1.46          & 45.12±0.82          \\
			\textbf{SD-DARTS }                                                               & \textbf{91.21±0.11}               & \textbf{93.91±0.08 }              & 70.87±0.33             & \textbf{71.88±0.76}               & \textbf{45.75±0.26}           & \textbf{46.03±0.59}               \\ \hline
		\end{tabular}
	\end{center}
	\vspace{-12pt}
\end{table}

\subsection{Results on NAS-Bench-201 Search Space}

NAS-Bench-201 \cite{2020arXiv200100326D} is the most widely used NAS benchmark. NAS-Bench-201 contains  4 internal nodes with 5 candidate operations. The search space includes 15,625 architectures, and the ground truth performance of each architecture on CIFAR-10, CIFAR-100 and ImageNet16-120 is provided. The searching settings of our method are the same as DARTS on on NAS-Bench-201.

Table \ref{201} shows comparison results on NAS-Bench-201. We  search on CIFAR-10 and use the found genotype to
query the performance of various datasets. It is observed that our method achieves competitive results on CIFAR-10, CIFAR-100, and ImageNet16-120 when compared to other NAS methods. Although the CIFAR-100-valid performance of our method is slightly lower than DARTS- \cite{chu2020darts}, it outperforms other methods such as PC-DARTS \cite{xu2019pc} and iDARTS \cite{pmlr-v139-zhang21s} in terms of overall performance. These results indicate that SD-DARTS has the capability to discover architectures with competitive performance across different datasets. While it may not achieve the absolute state-of-the-art on every individual dataset, it offers a strong trade-off between search efficiency and performance. These results highlight the value of our method in finding high-quality architectures for a wide range of applications. 

\begin{figure}[h]
	\vskip 0.2in
	\begin{center}
		\centerline{\includegraphics[width=10cm]{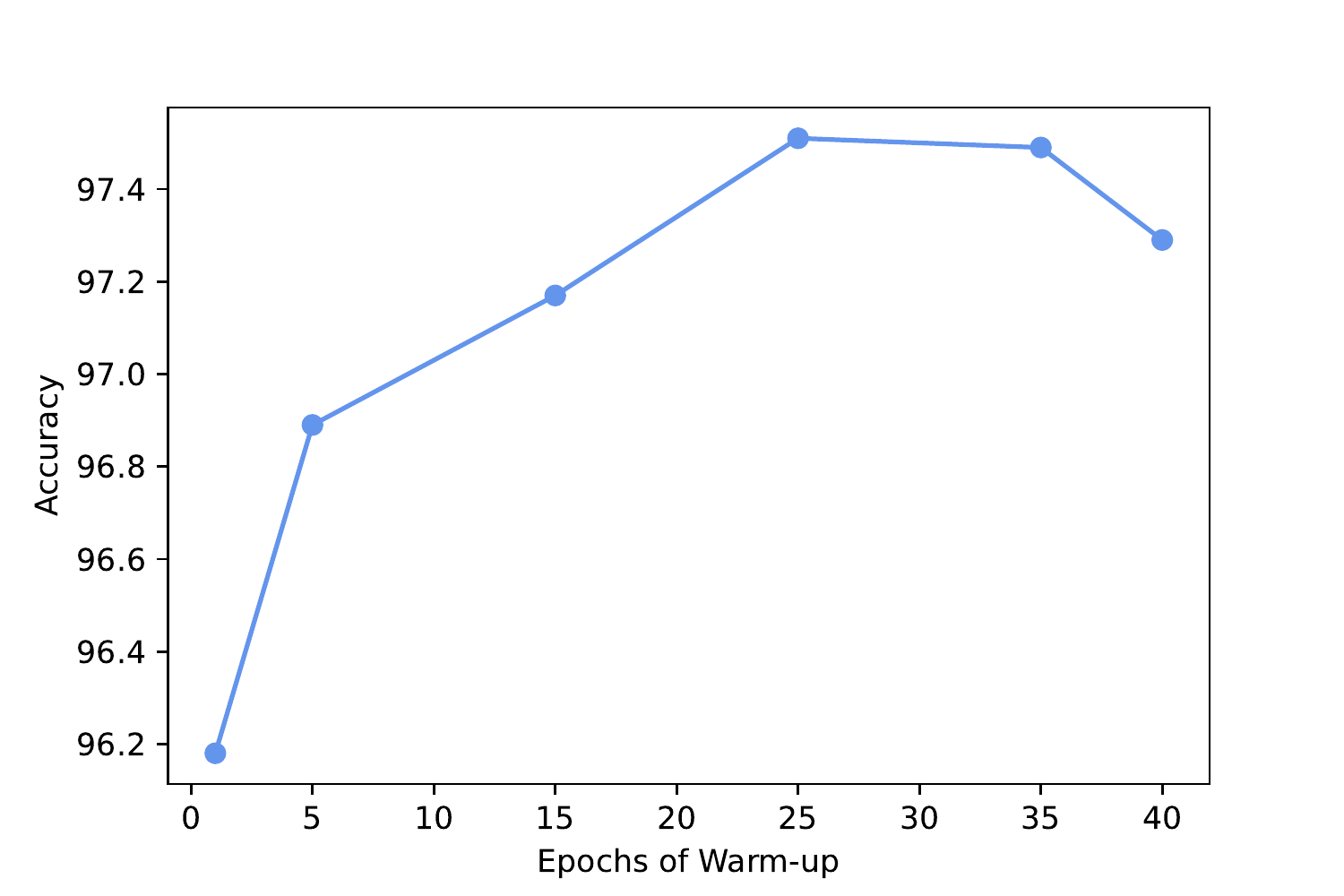}}
		\caption{The effect of warm-up : accuracy of the architectures discovered by SD-DARTS in different warm-up epochs on CIFAR-10. The accuracy grows gradually when the warm-up epochs increase until 25.}
		\label{warm-up}
	\end{center}
	\vskip -0.2in
\end{figure}

\section{Ablation Study}

\subsection{The Effect of Warm-up}

\label{ewp}
Because architecture and network parameters are randomly initialized, we need to warm up the supernet. After the warm-up, the supernet learns enough knowledge from train data to become a qualified teacher. In the subsection, we make an ablation study about the importance of warm-up.

We set  time window $K$ as 1 in the experiment, i.e., we utilize the supernet at the previous time step as the teacher to guide the training of the supernet. Our aim to choose a single teacher is to neglect the effect of multiple teachers. We run our method several times with different epochs for warm-up. Then, we evaluate the architecture's performance discovered by our experiments.

Fig. \ref{warm-up} shows the tendency of the accuracy of the architectures discovered by SD-DARTS in the different warm-ups. We find that the accuracy grows gradually when the warm-up epochs increase until 25. When the warm-up epochs are insufficient, the supernet cannot learn enough knowledge to become a qualified teacher. It will misguide the training of supernet and generate architecture with poor performance. When the warm-up epochs increase from 25 epochs, the accuracy decreases gradually. This is because the supernet is over-optimized as the training epochs increase, and the supernet plays a negative role when it is taken for a teacher. Thus, warm-up plays an essential role in our method, and we are supposed to warm up the supernet for proper epochs.

\subsection{The Effect of Voting Teachers}
\label{eovts}

\begin{figure}[h]
	\vskip 0.2in
	\begin{center}
		\centerline{\includegraphics[width=10cm]{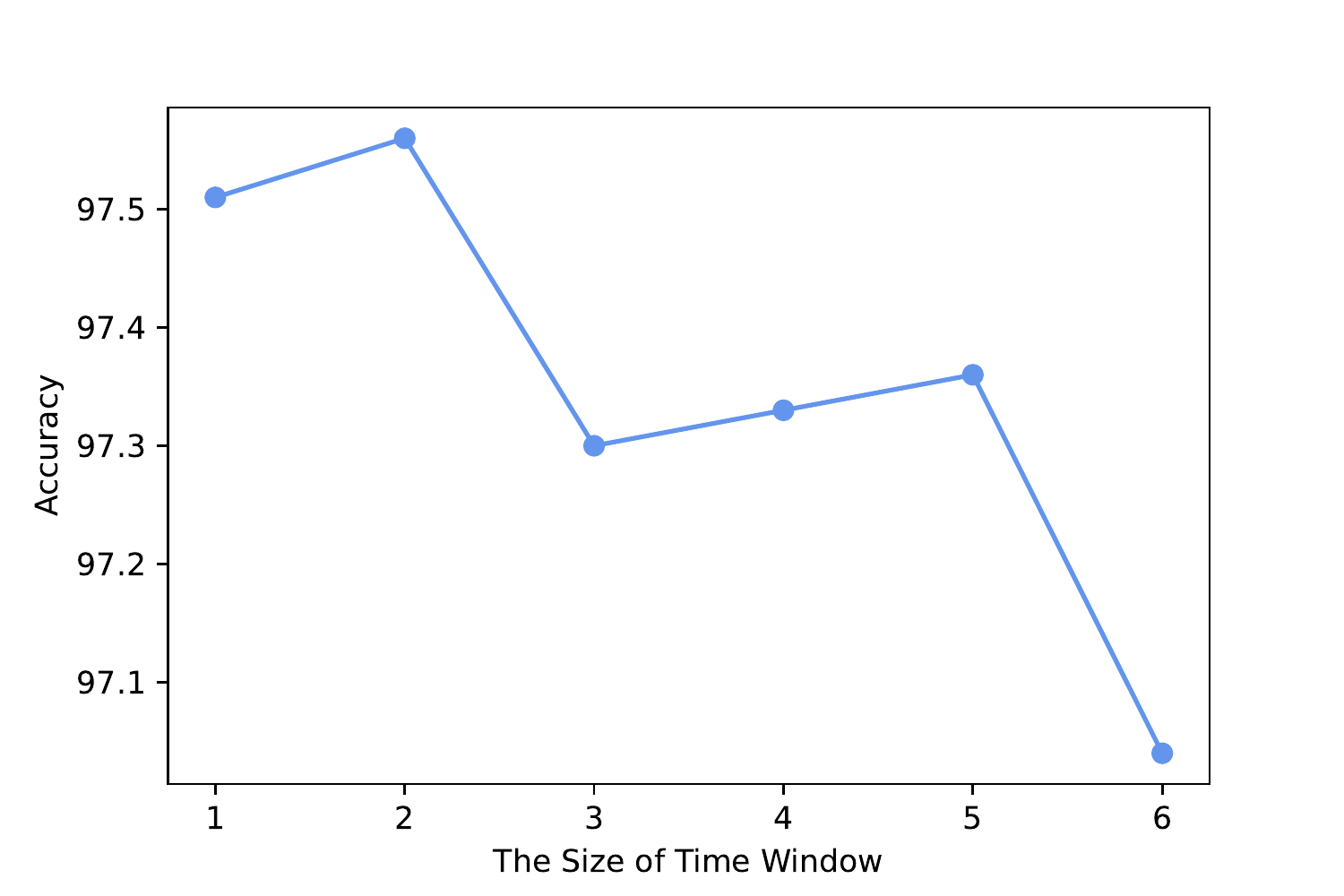}}
		\caption{The effect of voting teachers : accuracy of the architectures discovered by SD-DARTS in different time windows on CIFAR-10 when the epochs of warm-up are 25. When the size of time window is 2, it can achieve the highest accuracy. However, as the size of time window gets larger, the accuracy  declines slowly.}
		\label{time-window}
	\end{center}
	\vskip -0.2in
\end{figure}


In this subsection, the effect of voting teachers is analyzed by considering different time windows $K$, where $K$ represents the number of voting teachers. The warm-up epochs are set to 25, and the optimal architecture is searched using several supernets from the previous $K$ time steps as teachers to guide the self-distillation training of the supernet. The performance of these architecture is then evaluated. The purpose of this analysis is to understand the impact of the number of voting teachers on the performance of the search process. By varying the value of $K$, different amounts of knowledge and information from previous supernets are integrated to guide the training. This allows for a comprehensive exploration of the effect of using multiple teachers in the self-distillation process. 

%

Fig. \ref{time-window} illustrates the tendency of accuracy in different time windows when using the voting teachers method. It shows that the performance of the searched architectures varies with the size of the time window. Interestingly, the best performance is achieved when the time window is set to 2, indicating that utilizing the supernets from the previous two time steps as the teachers provides the most effective guidance for the self-distillation training of the supernet. As the size of the time window increases beyond 2, the performance of the architectures starts to deteriorate. This can be attributed to the fact that the information from supernets in earlier time steps becomes less relevant and correlated with the supernet at the present time step. Consequently, using supernets from these earlier time steps as voting teachers may introduce misleading information and hinder the effectiveness of the self-distillation training. This observation highlights the trade-off between the number of voting teachers and the correlation between the teachers and the student. Future research can focus on finding a balance between these factors to fully leverage the information from supernets at different time steps and optimize the guidance provided by the voting teachers in the self-distillation process. This would allow for a more effective and accurate search for optimal architectures.

\section{Conclusion}

In summary, the proposed SD-DARTS method, which combines self-distillation and voting teachers, aims to improve the performance of DARTS by addressing the sharpness of the loss landscape. It achieves state-of-the-art performance on CIFAR-10 and ImageNet datasets by reducing the discretization gap between the supernet and the optimal subnet. While SD-DARTS demonstrates promising results, it also has limitations regarding the utilization of information from supernets at different time steps to guide the training of the supernet. In future research, it would be valuable to explore alternative approaches, such as using a decayed exponential average, to strike a better balance between the number of voting teachers and the correlation between the teachers and the student. This could potentially enhance the performance and effectiveness of SD-DARTS even further.




\section*{Acknowledgments}

The work of Jian Li is supported partially by Natural Science Foundation of China (No. 62106257), China Postdoctoral Science Foundation (No. 2023T160680), and Excellent Talents Program of Institute of Information Engineering, CAS.
The work of Yong Liu is supported partially by Natural Science Foundation of China (No. 62076234), Beijing Outstanding Young Scientist Program (No. BJJWZYJH012019100020098), the Unicom Innovation Ecological Cooperation Plan, and the CCF-Huawei Populus Grove Fund. 

\bibliographystyle{elsarticle-num}
\bibliography{ref}{}

\begin{thebibliography}{10}
\expandafter\ifx\csname url\endcsname\relax
  \def\url#1{\texttt{#1}}\fi
\expandafter\ifx\csname urlprefix\endcsname\relax\def\urlprefix{URL }\fi
\expandafter\ifx\csname href\endcsname\relax
  \def\href#1#2{#2} \def\path#1{#1}\fi

\bibitem{deng2009imagenet}
J.~Deng, W.~Dong, R.~Socher, L.-J. Li, K.~Li, L.~Fei-Fei, Imagenet: A
  large-scale hierarchical image database, in: IEEE conference on computer
  vision and pattern recognition (CVPR), 2009, pp. 248--255.

\bibitem{yilmaz2006object}
A.~Yilmaz, O.~Javed, M.~Shah, Object tracking: A survey, Acm computing surveys
  (CSUR) 38~(4) (2006) 13--es.

\bibitem{van1996new}
M.~van Heel, G.~Harauz, E.~V. Orlova, R.~Schmidt, M.~Schatz, A new generation
  of the imagic image processing system, Journal of structural biology 116~(1)
  (1996) 17--24.

\bibitem{li2000automatic}
H.~Li, D.~Doermann, O.~Kia, Automatic text detection and tracking in digital
  video, IEEE transactions on image processing 9~(1) (2000) 147--156.

\bibitem{simonyan2014very}
K.~Simonyan, A.~Zisserman, Very deep convolutional networks for large-scale
  image recognition, in: Y.~Bengio, Y.~LeCun (Eds.), International Conference
  on Learning Representations (ICLR), 2015.

\bibitem{he2016deep}
K.~He, X.~Zhang, S.~Ren, J.~Sun, Deep residual learning for image recognition,
  in: Proceedings of the IEEE conference on computer vision and pattern
  recognition (CVPR), 2016, pp. 770--778.

\bibitem{vaswani2017attention}
A.~Vaswani, N.~Shazeer, N.~Parmar, J.~Uszkoreit, L.~Jones, A.~N. Gomez,
  {\L}.~Kaiser, I.~Polosukhin, Attention is all you need, Advances in neural
  information processing systems (NeurIPS) 30 (2017).

\bibitem{devlin2018bert}
J.~Devlin, M.~Chang, K.~Lee, K.~Toutanova, {BERT:} pre-training of deep
  bidirectional transformers for language understanding, in: J.~Burstein,
  C.~Doran, T.~Solorio (Eds.), Proceedings of the Conference of the North
  American Chapter of the Association for Computational Linguistics (NAACL),
  2019, pp. 4171--4186.

\bibitem{brown2020language}
T.~Brown, B.~Mann, N.~Ryder, M.~Subbiah, J.~D. Kaplan, P.~Dhariwal,
  A.~Neelakantan, P.~Shyam, G.~Sastry, A.~Askell, et~al., Language models are
  few-shot learners, Advances in neural information processing systems
  (NeurIPS) 33 (2020) 1877--1901.

\bibitem{zoph2016neural}
B.~Zoph, Q.~V. Le, Neural architecture search with reinforcement learning, in:
  International Conference on Learning Representations (ICLR), 2017.

\bibitem{real2019regularized}
E.~Real, A.~Aggarwal, Y.~Huang, Q.~V. Le, Regularized evolution for image
  classifier architecture search, in: Proceedings of the aaai conference on
  artificial intelligence (AAAI), Vol.~33, 2019, pp. 4780--4789.

\bibitem{ostad2021subsurface}
K.~Ostad-Ali-Askari, M.~Shayan, Subsurface drain spacing in the unsteady
  conditions by hydrus-3d and artificial neural networks, Arabian Journal of
  Geosciences 14 (2021) 1--14.

\bibitem{zoph2018learning}
B.~Zoph, V.~Vasudevan, J.~Shlens, Q.~V. Le, Learning transferable architectures
  for scalable image recognition, in: Proceedings of the IEEE conference on
  computer vision and pattern recognition (CVPR), 2018, pp. 8697--8710.

\bibitem{ostad2017artificial}
K.~Ostad-Ali-Askari, M.~Shayannejad, H.~Ghorbanizadeh-Kharazi, Artificial
  neural network for modeling nitrate pollution of groundwater in marginal area
  of zayandeh-rood river, isfahan, iran, KSCE Journal of Civil Engineering 21
  (2017) 134--140.

\bibitem{pham2018efficient}
H.~Pham, M.~Guan, B.~Zoph, Q.~Le, J.~Dean, Efficient neural architecture search
  via parameters sharing, in: International conference on machine learning
  (ICML), 2018, pp. 4095--4104.

\bibitem{DBLP:conf/eccv/GuoZMHLWS20}
Z.~Guo, X.~Zhang, H.~Mu, W.~Heng, Z.~Liu, Y.~Wei, J.~Sun, Single path one-shot
  neural architecture search with uniform sampling, in: A.~Vedaldi, H.~Bischof,
  T.~Brox, J.~Frahm (Eds.), European Conference on Computer Vision (ECCV), Vol.
  12361, 2020, pp. 544--560.

\bibitem{luo2018neural}
R.~Luo, F.~Tian, T.~Qin, E.~Chen, T.-Y. Liu, Neural architecture optimization,
  Advances in neural information processing systems (NeurIPS) 31 (2018).

\bibitem{xu2019pc}
Y.~Xu, L.~Xie, X.~Zhang, X.~Chen, G.~Qi, Q.~Tian, H.~Xiong, {PC-DARTS:} partial
  channel connections for memory-efficient architecture search, in:
  International Conference on Learning Representations (ICLR), 2020.

\bibitem{li2020sgas}
G.~Li, G.~Qian, I.~C. Delgadillo, M.~Muller, A.~Thabet, B.~Ghanem, Sgas:
  Sequential greedy architecture search, in: Proceedings of the IEEE/CVF
  Conference on Computer Vision and Pattern Recognition (CVPR), 2020, pp.
  1620--1630.

\bibitem{zela2019understanding}
A.~Zela, T.~Elsken, T.~Saikia, Y.~Marrakchi, T.~Brox, F.~Hutter, Understanding
  and robustifying differentiable architecture search, in: International
  Conference on Learning Representations (ICLR), 2020.

\bibitem{chen2020stabilizing}
X.~Chen, C.-J. Hsieh, Stabilizing differentiable architecture search via
  perturbation-based regularization, in: International conference on machine
  learning (ICML), 2020, pp. 1554--1565.

\bibitem{DBLP:journals/ijcv/ChenXWT21}
X.~Chen, L.~Xie, J.~Wu, Q.~Tian, Progressive {DARTS:} bridging the optimization
  gap for {NAS} in the wild, Int. J. Comput. Vis. 129~(3) (2021) 638--655.

\bibitem{https://doi.org/10.48550/arxiv.1909.06035}
H.~Liang, S.~Zhang, J.~Sun, X.~He, W.~Huang, K.~Zhuang, Z.~Li, {DARTS+:}
  improved differentiable architecture search with early stopping, CoRR
  abs/1909.06035 (2019).

\bibitem{wang2021rethinking}
R.~Wang, M.~Cheng, X.~Chen, X.~Tang, C.~Hsieh, Rethinking architecture
  selection in differentiable {NAS}, in: International Conference on Learning
  Representations (ICLR), 2021.

\bibitem{tian2021discretization}
Y.~Tian, C.~Liu, L.~Xie, Q.~Ye, et~al., Discretization-aware architecture
  search, Pattern Recognition 120 (2021) 108186.

\bibitem{bi2020gold}
K.~Bi, L.~Xie, X.~Chen, L.~Wei, Q.~Tian, {GOLD-NAS:} gradual, one-level,
  differentiable, CoRR abs/2007.03331 (2020).

\bibitem{chu2020darts}
X.~Chu, X.~Wang, B.~Zhang, S.~Lu, X.~Wei, J.~Yan, {DARTS-:} robustly stepping
  out of performance collapse without indicators, in: International Conference
  on Learning Representations (ICLR), 2021.

\bibitem{hinton2015distilling}
Y.~Lin, C.~Wang, C.~Chang, H.~Sun, An efficient framework for counting
  pedestrians crossing a line using low-cost devices: the benefits of
  distilling the knowledge in a neural network, Multim. Tools Appl. 80~(3)
  (2021) 4037--4051.

\bibitem{zhang2019your}
L.~Zhang, J.~Song, A.~Gao, J.~Chen, C.~Bao, K.~Ma, Be your own teacher: Improve
  the performance of convolutional neural networks via self distillation, in:
  Proceedings of the IEEE/CVF International Conference on Computer Vision
  (ICCV), 2019, pp. 3713--3722.

\bibitem{yun2020regularizing}
S.~Yun, J.~Park, K.~Lee, J.~Shin, Regularizing class-wise predictions via
  self-knowledge distillation, in: Proceedings of the IEEE/CVF conference on
  computer vision and pattern recognition (CVPR), 2020, pp. 13876--13885.

\bibitem{tarvainen2017mean}
A.~Tarvainen, H.~Valpola, Mean teachers are better role models: Weight-averaged
  consistency targets improve semi-supervised deep learning results, Advances
  in neural information processing systems (NeurIPS) 30 (2017).

\bibitem{li2020dual}
K.~Li, S.~Wang, L.~Yu, P.~A. Heng, Dual-teacher++: Exploiting intra-domain and
  inter-domain knowledge with reliable transfer for cardiac segmentation, IEEE
  Transactions on Medical Imaging 40~(10) (2020) 2771--2782.

\bibitem{zhao2022uda}
Z.~Zhao, F.~Zhou, K.~Xu, Z.~Zeng, C.~Guan, S.~K. Zhou, Le-uda: Label-efficient
  unsupervised domain adaptation for medical image segmentation, IEEE
  Transactions on Medical Imaging 42~(3) (2022) 633--646.

\bibitem{li2021hierarchical}
S.~Li, Z.~Zhao, K.~Xu, Z.~Zeng, C.~Guan, Hierarchical consistency regularized
  mean teacher for semi-supervised 3d left atrium segmentation, in:
  International Conference of the IEEE Engineering in Medicine \& Biology
  Society (EMBC), IEEE, 2021, pp. 3395--3398.

\bibitem{zhao2022act}
Z.~Zhao, A.~Zhu, Z.~Zeng, B.~Veeravalli, C.~Guan, Act-net: Asymmetric
  co-teacher network for semi-supervised memory-efficient medical image
  segmentation, in: IEEE International Conference on Image Processing (ICIP),
  2022, pp. 1426--1430.

\bibitem{wei2019online}
H.-R. Wei, S.~Huang, R.~Wang, X.~Dai, J.~Chen, Online distilling from
  checkpoints for neural machine translation, in: Proceedings of the Conference
  of the North American Chapter of the Association for Computational
  Linguistics (NAACL), 2019, pp. 1932--1941.

\bibitem{foret2021sharpnessaware}
P.~Foret, A.~Kleiner, H.~Mobahi, B.~Neyshabur, Sharpness-aware minimization for
  efficiently improving generalization, in: International Conference on
  Learning Representations (ICLR), 2021.

\bibitem{pmlr-v139-kwon21b}
J.~Kwon, J.~Kim, H.~Park, I.~K. Choi, Asam: Adaptive sharpness-aware
  minimization for scale-invariant learning of deep neural networks, in:
  Proceedings of the 38th International Conference on Machine Learning (ICML),
  Vol. 139, 2021, pp. 5905--5914.

\bibitem{2022arXiv220203599Z}
Y.~Zhao, H.~Zhang, X.~Hu, Penalizing gradient norm for efficiently improving
  generalization in deep learning, in: International Conference on Machine
  Learning (ICML), Vol. 162, 2022, pp. 26982--26992.

\bibitem{du2022sharpness}
J.~Du, D.~Zhou, J.~Feng, V.~Tan, J.~T. Zhou, Sharpness-aware training for free,
  in: NeurIPS, 2022.

\bibitem{liu2018darts}
H.~Liu, K.~Simonyan, Y.~Yang, {DARTS}: Differentiable architecture search, in:
  International Conference on Learning Representations (ICLR), 2019.

\bibitem{krizhevsky2009learning}
A.~Krizhevsky, G.~Hinton, et~al., Learning multiple layers of features from
  tiny images (2009).

\bibitem{huang2017densely}
G.~Huang, Z.~Liu, L.~Van Der~Maaten, K.~Q. Weinberger, Densely connected
  convolutional networks, in: Proceedings of the IEEE conference on computer
  vision and pattern recognition (CVPR), 2017, pp. 4700--4708.

\bibitem{liu2018progressive}
C.~Liu, B.~Zoph, M.~Neumann, J.~Shlens, W.~Hua, L.-J. Li, L.~Fei-Fei,
  A.~Yuille, J.~Huang, K.~Murphy, Progressive neural architecture search, in:
  Proceedings of the European conference on computer vision (ECCV), 2018, pp.
  19--34.

\bibitem{xie2018snas}
S.~Xie, H.~Zheng, C.~Liu, L.~Lin, {SNAS:} stochastic neural architecture
  search, in: International Conference on Learning Representations (ICLR),
  2019.

\bibitem{cai2018proxylessnas}
H.~Cai, L.~Zhu, S.~Han, Proxylessnas: Direct neural architecture search on
  target task and hardware, in: International Conference on Learning
  Representations (ICLR), 2019.

\bibitem{2022arXiv220301665Y}
P.~Ye, B.~Li, Y.~Li, T.~Chen, J.~Fan, W.~Ouyang, {\(\beta\)}-darts: Beta-decay
  regularization for differentiable architecture search, in: {IEEE/CVF}
  Conference on Computer Vision and Pattern Recognition (CVPR), 2022, pp.
  10864--10873.

\bibitem{dong2019searching}
X.~Dong, Y.~Yang, Searching for a robust neural architecture in four gpu hours,
  in: Proceedings of the IEEE/CVF Conference on Computer Vision and Pattern
  Recognition (CVPR), 2019, pp. 1761--1770.

\bibitem{huang2023u}
L.~Huang, S.~Sun, J.~Zeng, W.~Wang, W.~Pang, K.~Wang, {U-DARTS:} uniform-space
  differentiable architecture search, Inf. Sci. 628 (2023) 339--349.

\bibitem{li2023darts}
Y.~Li, S.~Li, Z.~Yu, {DARTS-PAP:} differentiable neural architecture search by
  polarization of instance complexity weighted architecture parameters, in:
  MultiMedia Modeling (MMM), Vol. 13834, 2023, pp. 277--288.

\bibitem{szegedy2015going}
C.~Szegedy, W.~Liu, Y.~Jia, P.~Sermanet, S.~Reed, D.~Anguelov, D.~Erhan,
  V.~Vanhoucke, A.~Rabinovich, Going deeper with convolutions, in: Proceedings
  of the IEEE conference on computer vision and pattern recognition (CVPR),
  2015, pp. 1--9.

\bibitem{howard2017mobilenets}
A.~G. Howard, M.~Zhu, B.~Chen, D.~Kalenichenko, W.~Wang, T.~Weyand,
  M.~Andreetto, H.~Adam, Mobilenets: Efficient convolutional neural networks
  for mobile vision applications, CoRR abs/1704.04861 (2017).

\bibitem{zhang2018shufflenet}
X.~Zhang, X.~Zhou, M.~Lin, J.~Sun, Shufflenet: An extremely efficient
  convolutional neural network for mobile devices, in: Proceedings of the IEEE
  conference on computer vision and pattern recognition (CVPR), 2018, pp.
  6848--6856.

\bibitem{ma2018shufflenet}
N.~Ma, X.~Zhang, H.-T. Zheng, J.~Sun, Shufflenet v2: Practical guidelines for
  efficient cnn architecture design, in: Proceedings of the European conference
  on computer vision (ECCV), 2018, pp. 116--131.

\bibitem{1999}
Y.~Liu, X.~Jia, M.~Tan, R.~Vemulapalli, Y.~Zhu, B.~Green, X.~Wang, Search to
  distill: Pearls are everywhere but not the eyes, in: 2020 {IEEE/CVF}
  Conference on Computer Vision and Pattern Recognition, {CVPR}, 2020, pp.
  7536--7545.

\bibitem{chu2021fairnas}
X.~Chu, B.~Zhang, R.~Xu, Fairnas: Rethinking evaluation fairness of weight
  sharing neural architecture search, in: Proceedings of the IEEE/CVF
  International Conference on Computer Vision (ICCV), 2021, pp. 12239--12248.

\bibitem{tan2019mnasnet}
M.~Tan, B.~Chen, R.~Pang, V.~Vasudevan, M.~Sandler, A.~Howard, Q.~V. Le,
  Mnasnet: Platform-aware neural architecture search for mobile, in:
  Proceedings of the IEEE/CVF Conference on Computer Vision and Pattern
  Recognition (CVPR), 2019, pp. 2820--2828.

\bibitem{russakovsky2015imagenet}
O.~Russakovsky, J.~Deng, H.~Su, J.~Krause, S.~Satheesh, S.~Ma, Z.~Huang,
  A.~Karpathy, A.~Khosla, M.~Bernstein, et~al., Imagenet large scale visual
  recognition challenge, International journal of computer vision 115~(3)
  (2015) 211--252.

\bibitem{2020arXiv200100326D}
X.~Dong, Y.~Yang, Nas-bench-201: Extending the scope of reproducible neural
  architecture search, in: International Conference on Learning Representations
  (ICLR), 2020.

\bibitem{Hu_2020_CVPR}
S.~Hu, S.~Xie, H.~Zheng, C.~Liu, J.~Shi, X.~Liu, D.~Lin, Dsnas: Direct neural
  architecture search without parameter retraining, in: Proceedings of the
  IEEE/CVF Conference on Computer Vision and Pattern Recognition (CVPR), 2020.

\bibitem{pmlr-v139-zhang21s}
M.~Zhang, S.~W. Su, S.~Pan, X.~Chang, E.~M. Abbasnejad, R.~Haffari, idarts:
  Differentiable architecture search with stochastic implicit gradients, in:
  Proceedings of the 38th International Conference on Machine Learning (ICML),
  Vol. 139, 2021, pp. 12557--12566.

\bibitem{chu2021darts}
X.~Chu, X.~Wang, B.~Zhang, S.~Lu, X.~Wei, J.~Yan, {\{}DARTS{\}}-: Robustly
  stepping out of performance collapse without indicators, in: International
  Conference on Learning Representations (ICLR), 2021.

\end{thebibliography}

\end{document}